# Real-Time and Efficient Method for Accuracy Enhancement of Edge Based License Plate Recognition System

Reza Azad*, Babak Azad**, Hamid Reza Shayegh Brojeeni***
* Shahid Rajaee Teacher Training University, Tehran , Iran, rezazad68@gmail.com
** Shahid Bahonar college, Shiraz , Iran, babak.babi72@gmail.com
*** Assistant professor, Shahid Rajaee Teacher Training University, Tehran,Iran, hamid.Shayegh@gmail.com

*Abstract:* License Plate Recognition plays an important role on the traffic monitoring and parking management. Administration and restriction of those transportation tools for their better service becomes very essential. In this paper, a fast and real time method has an appropriate application to find plates that the plat has tilt and the picture quality is poor. In the proposed method, at the beginning, the image is converted into binary mode with use of adaptive threshold. And with use of edge detection and morphology operation, plate number location has been specified and if the plat has tilt; its tilt is removed away. Then its characters are distinguished using image processing techniques. Finally, K Nearest Neighbour (KNN) classifier was used for character recognition. This method has been tested on available data set that has different images of the background, considering distance, and angel of view so that the correct extraction rate of plate reached at 98% and character recognition rate achieved at 99.12%. Further we tested our character recognition stage on Persian vehicle data set and we achieved 99% correct recognition rate.

## 1. Introduction

Vehicle License Plate Recognition (VLPR) also known as Automatic Vehicle License Plate Recognition (AVLPR) was invented in 1976. Many scientist groups took interest in VLPR after 1990s with the development of digital camera and the increase in processing speed. VLPR is an image processing technology which enables to extract vehicle license number form digital images. It consists of a still or video camera which takes the image of vehicle, find the location of the number in the image and then segments the characters and by using the k -Nearest Neighbour (KNN) scheme, it translates the license number of pixel value into numerical or string. VLPR can be used in many areas such as parking inventory [1], [2], security control of restricted areas [3], traffic law enforcement [4]-[6], congestion pricing [7], and automatic toll collection [8].

Typical VLPR System consists of four modules: image acquisition, license plate extraction, character segmentation, and character recognition [9]. The efficiency and accuracy of the system largely depends on the second module and various approaches have been used for this purpose. To detect the region of car license plate, many techniques have been used. The algorithm presented in [10] using projection and Euclidean distance reaches 87% as its performance. The algorithm presented in [11] using sliding concentric windows and probabilistic NN reaches 86% as its overall performance. What is reported in [12] using filtering and template matching depicted its performance as 91%. The method presented in [13] with Gabor filter and connected component reported its plate detection rate as 91.7%. The algorithm reported in [14] applying edge analysis and feed forward NN reaches 92.3% as its character recognition rate.

In [15] adaptive boosting (AdaBoost) is combined with Haar-like features to obtain cascade classifiers for license plate extraction. The Haar-like features are commonly used for object detection. Using the Haar-like features makes the classifier invariant to the brightness, colour, size, and position of license plates. In [16], a new and fast vertical edge detection algorithm (VEDA) was proposed for license plate extraction. VEDA showed that it is faster than Sobel operator by about seven to nine times. In [17] and [18] combination of edge statistics and mathematical morphology showed very good results, but it is time consuming and because of this problem, [19] uses block-base algorithm.

In [20] a novel method called "N row distance" is implemented. This method scans an image with N row distance and counts the existent edges. If the number of the edges is greater than a threshold then the license plate is recognized, if not threshold have to be reduced and algorithm will be repeated. This method is fast and has good results for simple images. Disadvantage of this paper is that the edge based algorithms are sensitive to unwanted edges such as noise edges, and they fail when they are applied to complex images.

A wavelet transform-based algorithm is used in [21] for extraction of the important features to be used for license plate location. This method can locate more than one license plate in an image. Methods which are symmetry based are mentioned in [22]. In [23], firstly, it takes the input image into a grayscale, then for analyzing the location of plate the operation of morphology such as erosion and dilation is applied, and the plate is extracted with use of vertical and horizontal projection among various candidates.  In [24] the plate is a location with the black background and white writings. In this way that, firstly, takes the image into the HSI and applies the capability of being black colour of its background for this purpose, it uses a mask and segments the image according to HSI colour intensity parameter and creates a binary image. For cancelling probable noises, it uses the operation of erosion and dilation, then labels the existing candidates and for cancelling the candidates which aren't the location of plate, it applies the geometric capability of the plate and other characters, then for recognizing a primary candidate, it uses the colour intensity histogram, and recognizes the location of plate.

In this paper a new method based on mathematical morphology and KNN classifier is presented that in comparison with other methods has better operation on a condition that plate was tilt in the images or ambient light was low. In section 2, the proposed method for license plate recognition (LPR) and character recognition is elaborated, in section 3 the practical result of the paper and in section 4, conclusion is presented.

## 2. Proposed Method

General diagram of the proposed method is shown in Fig. 1. In the proposed method, the input image is reconstructed at the first time and then plate location is specified by edge detection and morphology operation and plate tilt is corrected and finally vehicle license plate location is extracted and their characters are recognized.

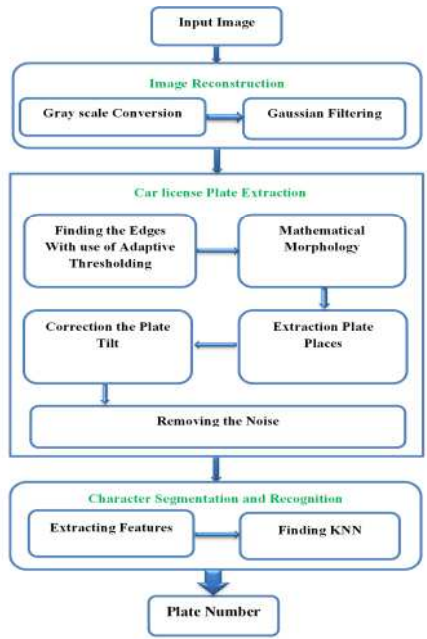

Fig. 1: General diagram of the proposed method

### 2.1 Image Reconstruction

Pre-processing is carried out on the image to improve the quality of the image so that the main processing on the image becomes easier. This step involves image converting to grey scale and Gaussian filter applying.

### 2.1.1 Gray Scale Converting

The license plate location and recognition in this paper are based on gray image, so the main function of the pretreatment algorithm is to convert colour images to gray scale images for the latter operation. A colour bitmap is composed of R, G and B 3 components. If it is a 24-bit true colour image, every point is made up of three bytes which respectively represent R, G and B. Therefore, the following is colour information of image. Equation (1) shows this converting.

$$\begin{cases} \forall I, J: \; I \in \text{Color.image(rows)} \text{ AND } J \in \text{Color.image(columns)} \\ \text{Gray.image}(I, J) = 0.299 * R(I, J) + 0.587 * G(I, J) + 0.114 * B(I, J) 1 \end{cases} \quad (1)$$

### 2.1.2 Gaussian Filtering

After converting the image into gray level, image is improved to remove the possible noise on it. To this end, a Gaussian filter with (2) equation is used.

$$f(x) = \exp\left(\frac{-0.5(x-c)^2}{\sigma^2}\right) \quad (2)$$

Where c is the mean and σ is the variance of gray image.

## 2.2 Exploiting of Plate Place

In this part, the candidate region has been identified with use of edge detection and mathematical morphology, and then among these areas, plate area is extracted.

### 2.2.1 Vertical Edge Detection

Edge detection is a type of image segmentation techniques which determines the presence of an edge or line in an image and outlines them in an appropriate way [25]. The main purpose of edge detection is to simplify the image data in order to minimize the amount of data to be processed [26]. Generally, an edge is defined as the boundary pixels that connect two separate regions with changing image amplitude attributes such as different constant luminance and tristimulus values in an image [25], [27], [28]. The number of vehicle due to written numbers and letters on it has many vertical edges that use these features to find plate location in the picture. There are different approaches and algorithm to find out the edge in image processing that, in the meantime, Sobel operator due to high speed and low processing volume has a more favorable performance compared to other methods. Sobel edge detection method has of both vertical and horizontal edge detection. In the proposed method both of them are used. Fig. 2(a) shows the Sobel vertical mask and Fig. 2(b) shows the Sobel horizontal mask.

| -1 | 0 | 1 |
|---|---|---|
| -2 | 0 | 2 |
| -1 | 0 | 1 |

| -1 | -2 | -1 |
|---|---|---|
| 0 | 0 | 0 |
| 1 | 2 | 1 |

Fig. 2: (a): Sobel vertical mask   (b): Sobel horizontal

Before applying this filter, first of all, the gray image is converted into binary image with use of statistical features. To this end, at first, the mean and variance of the image is calculated. Then the threshold level is placed equal to the sum of the mean and variance. And the level of the image that is more the threshold are converted into intensity values 1, and other values are converted into intensity values zero. The equation (3) shows the threshold applying way.

$$\begin{cases} \text{Mean(Gray)} = \frac{1}{n \times m} \sum_{1 \leq i \leq n} \sum_{1 \leq j \leq m} \text{Gray}(i,j) \\ \text{Var(Gray)} = \frac{1}{n \times m} \sum_{\substack{1 \leq i \leq m \\ 1 < j < n}} (\text{Gray}(i,j) - \text{Mean(Gray)})^2 \quad \rightarrow \text{Binary.image}[i,j] = \begin{cases} 1 \; if \; Gray(i,j) > Threshold \\ 0 \; if \; Gray(i,j) < Threshold \end{cases} \\ \text{Threshold} = \text{Mean(Gray.image)} + \text{Var(Gray.image)} \quad (3) \end{cases}$$

Where, Gray (i, j) is the value of pixel in ith row and jth column in Gray channel. Also, n and m are the size of image. Fig.3 shows these steps ceaselessly.

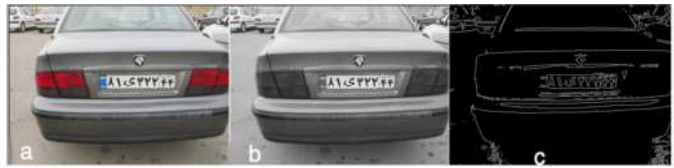

Fig. 3: (a): Original image (b): image in form of gray scale (c): Edge detected image

### 2.2.2 Mathematical Morphology

Mathematical morphology is the branch of image processing that argues about shape and appearance of abject in images. The erosion and dilation operators are basically operators of mathematical morphology that are used in this part to improve the edge detection image. At this step, first erosion action is applied in the edge detection image. The erosion action is defined as equation (4).

$$A \theta B = \{X | (B)_X \in A\} \quad (4)$$

After erosion action on image, the dilation action is done. The dilation action is defined by the equation (5).

$$A \theta B = \{X | ((B)_X \cap A) \in A\} \quad (5)$$

A and B dilation is the collection of all X that Bx and at least A have overlap in a non-zero element. Bx is B symmetric around its own axis. And then it transfers the symmetry of X. Since the plate area is a compact area. In this stage the holes are filled.

## 2.3 Extraction Plate Places

After morphology applied in the image, several interconnected areas will be produced. First, the length of each of these areas is obtained and an area that has the greatest length and is in the geometries range of plate and accepted by the density testing, is considered as the exact location of license plate and is placed inside a rectangle; then the rest area is eliminated. Following function show plate place extracting method.

---

Function Plate place Extraction(Candidate$_{(plate)}$)
(1): $\forall I, j: \left( I \in Candidate_{Plate(rows)} \text{ AND } j \in Candidate_{Plate(cloumns)} \right)$ Do
(2): Array[i] = $\sum color\_Jump(candidate_{plate(i,j)})$ End
(3): If There is Exist [$f(i,j): i, j \in$ Array AND $|j - i| > 12$ And $\forall i, j \in$ Range$(j - i)$: Value$(i, j) > 15$ )] Then
(4): Accept Candidate as Plate places.
(5): Else Candidate not plate places.
End if
End Function
Color_Jump(Candidate$_{Plate(i,j)}$) = Color_Jump(Candidate$_{Plate(i,j)}$) XOR Color_Jump(Candidate$_{Plate(i+1,j+1)}$)

After the candidate validated as a plate, it located in oblong and if it consist tilt it will be solved. Fig. 4(a) shows exploited plate from entrance image.

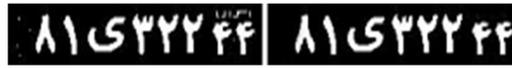

Fig. 4: (a): image of extracted plate (b): image of extracted plate after removing the noise

At this stage, license plate space can be an unequal figure that causes an error in the reading of the characters. So, since the license plate is located in a 4-point rectangle and offers the license plate corners information, this feature is used to remove the plate tilt.

### 2.3.1 Correction the Plate Tilt

In the process of the vehicle license plate recognition, the tilt of license plate has significant influence on the character segmentation, identification of patterns and the final recognition results [29]. So at this stage, first, by using the plate features that is determined by the candidate, the image of plate is extracted of original image; then it is improved in contrast and is converted into the binary level. To remove tilt of plate, the features of pixels arrangements is used in a digital image. To this end, in image without tilt, the path that goes from one corner to another corner of the image must include only pixels that are connected together from one corner. In this paper this method is used to remove the plate rotating that shows the good result about 45 degrees.

### 2.4 Character Segmentation and Recognition

To isolate the characters of car license plate, many techniques have been used. In [30], the extracted license plate is resized into a known template size. In this template, all character positions are known. After resizing, the same positions are extracted to be the characters. This method has the advantage of simplicity. However, in the case of any shift in the extracted license plate, the extraction results in background instead of characters. Since characters and license plate backgrounds have different colours, they have opposite binary values in the binary image. Therefore, some proposed methods as in [31]-[42] project the binary extracted license plate vertically to determine the starting and the ending positions of the characters, and then project the extracted characters horizontally to extract each character alone. Segmentation is performed in [43]–[48] by labelling the connected pixels in the binary license plate image. The labelled pixels are analysed and those which have the same size and aspect ratio of the characters are considered as license plate characters. This method fails to extract all the characters when there are joined or broken characters.

In this paper from last step, the area that is exploited as a plate, first probable noising are solved, then plate image is complemented till its writing of plate inside is seen such white violence. Then this area is labelled and through the available regions, the regions that are bigger are stored as exploited characters in 30*15 sizes. Fig. 4(b) shows the extracted plate after removing noise and Fig. h shows its histograms.

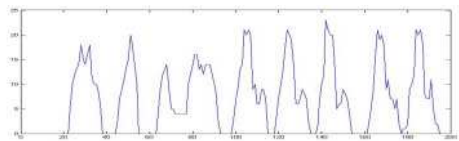

Fig. 5: Histogram of extracted plate

The extracted characters are then recognized and the output is the license plate number. To recognition the characters of car license plate, many techniques have been used. In [49], the feature vector is generated by dividing the binary character into blocks of 3×3 pixels. Then, the number of black pixels in each block is counted. In [50], the feature vector is generated by dividing the binary character after a thinning operation into 3 × 3 blocks and counting the number of elements that have 0°, 45°, 90°, and 135° inclination. In [51], the character is scanned along a central axis. This central axis is the connection between the upper bound horizontal central moment and lower bound horizontal central moment. Then the number of transitions from character to background and spacing between them form a feature vector for each character. This method is invariant to the rotation of the character because the same feature vector is generated. Template matching is performed in [52] - [55] after resizing the extracted character into the same size. Several similarity measuring techniques are defined in the literature. Some of them are Mahalanobis distance and the Bayes decision technique [53], Jaccard value [54], Hausdorff distance [55].

For recognition of characters in this paper, first characters which are obtained of former step in 60*30 sizes are thinned. Then any characters is broken to stable sizes that are called frame; two features are exploited from each frames, 1) mean of distance, which it is distance of any white pixel to corner of frame, 2) one angle which its size is equal with mean of angles size of any white pixel to horizontal level. These two features are exploited for every frame and all of them are stored in feature vector then are allude to their symbol by KNN classifier.

### 2.4.1. Thinning

In this part, goal is finding Skelton of any characters, without any rupture or erosion in Skelton of characters. Different algorithm has been presented for character thinning such as SPTA algorithm in [56]. In this paper the method that has been proposed in [57] is used. Fig. 6 shows sample of its action on image of exploited characters.

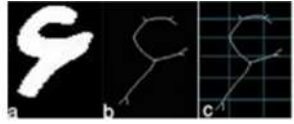

Fig. 6: (a): (a): Extracted character (b): The Skelton of the character (c): Framed character

For extracting of features, first image of character is broken to 18 frames in 10*10 sizes. This act is done by supposed window that is changes the 60*30 image to 6*3 supposed images. Then two features are extracted from these frames: 1) mean of distance, 2) angle average. Fig. 6(c) shows framing mood of characters.

### 2.4.2. Distance Feature

In this step distance mean of whole pixel of frame which pixel colour is white, are measured with pixel that are in left and bottom coroner of frame. With regard to number of frame, 18 features of distance mean are extracted for any image. Equation (6) shows this feature extraction:

$$(y_b) = \frac{1}{n} + \sum_{k=1}^{n_b} d_k^b \qquad , b=1,2,3,\ldots \qquad (6)$$

In top relation $y_b$ is distance mean for any frame and $d_k^b$ is distance of white pixel of agent of any frame.

### 2.4.3. Angle Feature

Angle features are very important features in order to achieve higher recognition accuracy and reducing misclassification. For achieve this feature mean of all pixel of frame which pixel colour is white, are measured with horizontal level of frame. With regard to number of frame, 18 features of angle mean are extracted for any image. Equation (7) shows this feature extraction:

$$(a_b) = \frac{1}{n} + \sum_{k=1}^{n_b} \theta_k^b \qquad , b=1,2,3,\ldots \qquad (7)$$

In the top relation, $a_b$ is angel average for any frame and $\theta_k^b$ angle of white pixel to frame horizontal level.

### 2.4.4. Classification

Classification stage uses the features extracted in the feature extraction stage for deciding the class membership. Classification phase is the decision making phase of a VLPR engine. In this work, we have used k-NN classifier for recognition. In the k-nearest neighbour classifier, Euclidean distances from the candidate vector to stored vector are computed. The Euclidean distance between a candidate vector and a stored vector is given by equation (8),

$$d = \sqrt{\sum_{k=1}^{N}(x_k - y_k)^2} \quad (8)$$

Here, N is the total number of features in feature set, $x_k$ is the library stored feature value and $y_k$ is the candidate feature value.

### 2.5. Determine Type of Plate

For specifying the kind of Iran country plates that are in three categories: 1) the public in yellow background, 2) the government in red background and 3) private cars in white background, are known. First plate image is converted to the HSV mode then the below algorithm is applied to recognized it.

Function Determining type of plate (plate)
(1): [n, m] ←size (plate)
(2): (∀i, j: i, j ∈ plate) Do {3, 4}
(3): X= $\begin{cases} \text{Red:} & \text{if } \{S \geq 0.45 \text{ And } V \geq 0.5 \text{ And } 0.8 \leq H \leq 0.94\} \\ \text{Yellow:} & \text{if } \{S \geq 0.45 \text{ And } V \geq 0.5 \text{ And } 0.58 \leq H \leq 0.74\} \\ \text{White:} & \text{f } \{S \leq 0.15 \quad\quad\quad \text{And } V \geq 0.8\} \end{cases}$
(4): Increase the Class(X) one time
(5): Kind=max (Class)
Show ("Type of Plate Color is (Kind)");
End Function

### 3. Practical Result

In this part first we will express the difference of threshold amount selection with adaptive state and stable state then proportion success rate of edge detection will be compered by Sobel operator and other operators that they are tested on different images practically. Also, the success rate of plate tilt solving that are tested on images, will express. Further the character recognition stage will be detailed. At the end the result of that has been done on the different images will express. Our suggestive method have been done on Intel Core i3-2330M CPU, 2.20 GHz with 2 GB RAM under Matlab environment.

### 3.1. Adaptive Thresholding

In this part threshold acts is shown with different amount for rising system efficiency on all images with different amount. The sample of practical result is shown on the image in fig. 7. Fig. 7(a) shows the state that the threshold amount is equal to 0.4. This amount because of lights may destroy some points in some images. It causes the plate faces with problem in recognizing step. Fig. 7(b) shows the state that the threshold amount is 0.7 in this situation extra points of image are made that they don't belong to plate area. Fig. 7(c) show the state that the threshold mound is selected by suggestive method so, in condition because the threshold mounts is selected image statistics information base; it expresses a better result in condition that the image light is less or more. Table 1 shows success rate, selection of different amount of threshold for finding plate place on our databases.

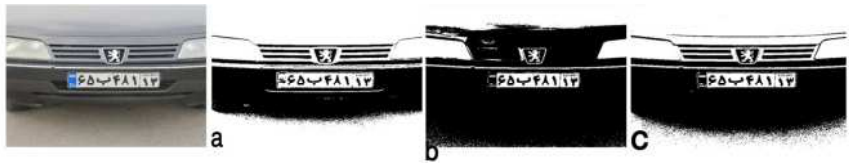

Fig. 7: a) Threshold with 0.4 value   b) Threshold with 0.7 value c) Adaptive Threshold

TABLE 1: Rate of success according to the threshold value

| Total image | Threshold value | Rate of success |
|---|---|---|
| 100 | ۰.۴ | %۴۰ |
|  | ۰.۷ | %۵۰ |
|  | adaptive | %۹۸ |

## 3.2. Edge Detection Method

There are different approaches and algorithm to find out the edge in image processing that, in the meantime, Sobel operator due to high speed and low processing volume has a more favorable performance compared to other methods. Table 2 shows the success rate of different methods for edge detection in our databases.

TABLE 2: Rate of success according to edge detection method

| Total image | Method | prewitt | Log | canny | Sobel | Robert | Zero cross |
|---|---|---|---|---|---|---|---|
| 100 | Success rate | 83% | 60% | 85% | 9^% | 88.22% | 70% |

## 3.3. Edge Detection Method

When the vehicle plate has tilted our suggestive method recognizes it and solves it. For solving plate tilt in this paper, the object orientation method is used. According to results that are done on all images, the images that the plate area is recognized correctly the success rate of this operator has been gained to 100%. Fig. 8 shows sample image that the plate has tilt but the system has known and solved it correctly.

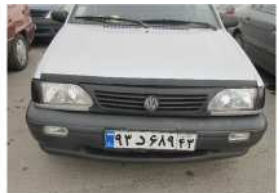

Fig. 8: car with tilt plate

### 3.4. Character Recognition

The results are based on two feature extraction techniques, namely, angle and distance features. We have also experimented some partitioning strategies. We have divided the character data set using three partitioning strategies. In the first strategy (strategy A), we have taken 70% data in training set and other 30% data in the testing set. In the second strategy (strategy B), we have considered 80% data in training set and remaining 20% data in the testing set. Strategy C has 100% data in training set and 100% data in testing set. These experimental results are depicted in Fig. 8.

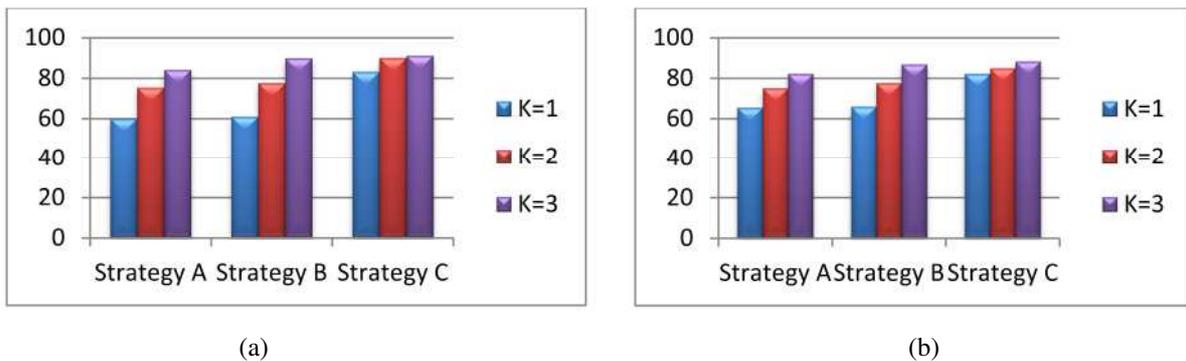

(a)          (b)
Fig. 8: a) Recognition accuracy with angle features using k-NN.    b) Recognition accuracy with distance features using k-NN.

Also The results of recognition accuracy with both angle and distance features are depicted in Table 3.

TABLE 3: Rate of character recognition based on both feature set

| Selecting Features | K value | Strategy A Accuracy | Strategy B Accuracy | Strategy C Accuracy |
|---|---|---|---|---|
| Angle and Distance | 2 | 99% | 99.12% | 100% |

### 3.4. Analysis of System

According to the previous license plate recognition (LPR) approaches, there is not a popular database of LPR to evaluate the performance of its methods. In this respect, in order to prove the quality of proposed approach, 100 images are captured by a 2 megapixel camera of mobile phone (Nokia 5230) from each view of car. These images are variety in point of view, various light condition, various distances and backgrounds. Next, the proposed approach is applied on them, and the accuracy rate is computed. Table4 shows the result.

TABLE 4: Performance of the Proposed Method

| Total image of car | Total characters | Correct license plate location | Correct character recognition | Percent Efficiency for each part | |
|---|---|---|---|---|---|
| 100 | 800 | 98 | 792 | 98% | 99.12% |

The cases which the system isn't able to recognize plate location it is because of background complexity. Further In our research experiment, we used [58] dataset that contains numerical and character images and all those image samples are extracted from the real world environment, such as parking lot, freeway and etc. for showing high accuracy of our method. In the table 5, our character recognition method is compared with normal factoring method represented in [59] and template matching mentioned in [58] that both of them used this dataset for evaluation of their works.

TABLE 5: Comparison rate of character recognition rate

| Total Image | Method | Technique | Correct character recognition | Percent Efficiency |
|---|---|---|---|---|
| **1200** | S.H.M Kasaei and etc. [58] | Template Maching | 1104 | 92% |
| | R.Azad and etc.[59] | Normal Factoring | 1164 | 97% |
| | R.Azad, H.R. Shayegh [Proposed Method] | Feature Set | 1188 | 99% |

In general, the advantages that are presented in this paper compared to other methods include the following: Lower computational complexity; fast response and operation; ability to correct plate tilt; ability to implementation on microprocessors; usability in real time work; detect minimum candidates as car plates. In most images, the proposed method detects a candidate that is in fact an original license plate; robust character recognition; scale invariant.

## 4. Conclusion

This paper suggested a quick method for license plate recognition. In this method, at first, with use of statistical features, an appropriate threshold will be obtained for the input image and converts it into the binary. Sober operator extracts the vertical edges of the image, then with use of morphology operation and the geometric ratio the location of plate is extracted and as result of this process, tilt of plate is set away so that when the system reads the characters no problem will ultimately arise. The proposed method has been examined on the image with different background, different distance and view point, various light and atmospheric conditions and also is in some situation when the image quality is low. What resulted from the examination of the method showed that the rate of correct extraction of the plate reached approximately at 98%, also character recognition stage obtained 99.12%. Further we tested our character recognition stage on Persian data set and we achieved high accuracy rate.